# A Survey of research in Deep Learning for Robotics for Undergraduate research interns

Narayanan PP, Maulana Azad National Institute of Technology

Palacode Narayana Iyer Anantharaman, JNResearch Labs LLP

**Disclaimer**

This document is a draft version at this stage and the final version will be created soon.

## 1. Abstract

Over the last several years use cases for robotics based solutions have diversified from factory floors to domestic applications. In parallel, Deep Learning approaches are replacing traditional techniques in Computer Vision, Natural Language Processing, Speech processing etc. and are delivering robust results. Our goal is to survey a number of research internship projects in the broad area of "Deep Learning as applied to Robotics" and present a concise view for the benefit of aspiring student interns. In this paper, we survey the research work done by Robotic Institute Summer Scholars (RISS), CMU. We particularly focus on papers that use deep learning to solve core robotic problems and also robotic solutions. We trust this would be useful particularly for internship aspirants for the Robotics Institute, CMU.

## 2. Introduction

The recent advances in Deep Learning, whether it be the evolution of deeper Convolutional Neural Networks (CNN) based models with hundreds of layers or the more recent large language models such as ChatGPT, have been phenomenal. One useful property of deep learning techniques is that the techniques are fairly generic and can be applied for solving problems in different domains. Hence it becomes a natural choice for robotic perception tasks besides many other novel applications where the deep learning models provide higher accuracy and robust results in the face of stochasticity.

Recently, robotics based use cases have diversified and robots are finding their way into our homes besides the traditional factory floor applications. For a survey of commercial home service robots see [21]. Deep learning enables many novel features and drives innovation in robotics. As an example, we can consider problems like Visual and Language Navigation (VLN) [26]. More recent examples of the intersection of deep learning with robotics include Google's Robotics Transformer 1 (RT-1) a multi-task model that tokenizes robot inputs and outputs actions (e.g., camera images, task instructions, and motor commands) to enable efficient inference at runtime, which makes real-time control feasible [28].

Our motivation behind this survey paper arose out of the need to understand the research reports/papers from RISS as a part of an internship application. We surveyed a number of

papers that are a subset of RISS papers published in 2021. We reviewed them using the following framework:

- What problem is being solved in this research?
- Is it generic enough to be applicable to a large set of applications?
- What is the solution proposed by the paper? What are the novel and innovative aspects?
- What kind of datasets are used? Are they available for download?
- Does this work provide adequate directions and opportunities to define other interesting problems that are suitable for aspiring undergraduate/graduate level research interns?

In the remaining sections we provide the main content of a sample of seven papers out of over 15 RISS papers we surveyed. The surveyed papers are chosen from deep learning for perception related tasks, task agnostic reinforcement learning, SLAM and Robotic applications.

## 3. Related Work

There are a number of survey papers that can be useful for understanding the current research in this field. We categorize them in to:

(i) Surveys pertaining to topics in Deep Learning, such as Computer Vision, Natural Language Processing, etc. For instance, the paper [29] by Zhuangdi Zhu et al. is a survey of transfer learning in Deep Reinforcement Learning.

(ii) Surveys focused on recent trends in Robotics Research particularly that describe the usage of deep learning to address robotics problems.

These surveys cover high end research in these topics and will be useful for senior researchers. These surveys are very useful for students to understand the broader research landscape but do not exclusively deal with internship level projects. Our focus is mainly the research conducted as a part of internship programs aimed at undergraduates. The motivation is to provide the necessary information that helps the student to identify his own internship problems, prepare himself/herself for an internship at top end research institutes.

## 4. RISS Survey

In this section we describe each of the papers from a sample of projects that we surveyed. Since these projects are done by different researchers with independent problem statements, there may not be a direct correlation between one project to the other though there may be commonality of the underlying techniques.

### 4.1 Class-Imbalanced Learning via Bilevel-Optimized Weight Decay

Refer [22]. Class imbalance refers to the scenario where the number of training samples belonging to some common classes are significantly higher than the number of training samples belonging to other classes. This can be a problem in machine learning because the classifier may become biased towards the classes with more samples, and it may be difficult for the classifier to learn the characteristics of rare classes. A naively trained classifier hence develops imbalanced norms of classifier weights, which results in common classes having large weights and the rare classes having small weights. As a result, the classifier may not perform well on the rare classes, leading to poor overall performance. To solve this problem, several studies [1] have focussed on developing Class-Imbalanced Learning(CIL) techniques with the aim to achieve higher accuracy averaged over imbalanced classes. Popular CIL methods emphasize rare-class performance through loss reweighting and data resampling.

In this paper, the authors propose a CIL algorithm to balance network weights via regularizing layer norms. This work specifically explores weight decay, or L2-normalization to learn small weights to overcome overfitting and presents a bilevel optimization technique to optimize weight decay parameters. Further, the experimentation results of this model suggests that the proposed technique outperforms existing CIL techniques measured using CIFAR 100 dataset.

Bilevel optimization is a type of optimization problem in which there are two levels of decision variables, with one level being optimized subject to the constraints of the other level. In model-free bilevel optimization, the decision variables at the lower level are not explicitly modeled. Instead, the lower level is treated as a "black box" and the optimization is performed directly on the output of the lower level. Prior works propose to use bilevel optimization to optimize hyperparameters along with learning model parameters. The typical criterion of optimizing hyperparameters is to maximize the performance on the validation set.

A bilevel optimization problem has two loops: the inner loop is training network parameters by minimizing errors on the training data, and the outer loop optimizes the hyperparameters by minimizing errors on the validation set. The weights and biases of the neural network typically represent the lower level variables while the hyperparameters of the model, such as learning rate $\alpha$ and regularization strength $\lambda$ represent higher level variables. The inner loop optimizes the lower level variables to minimize loss function and the outer loop optimizes the higher level variables to find a combination of hyperparameters that result in a well-performing neural network.

This work elaborates on the proposed "Population-based Bilevel Optimization(PBO)" method which is model-free. The essential idea of PBO is to 1. train a population of K models simultaneously using K different hyperparameters , 2. periodically evaluate K checkpoints,  3. select the best performing model on validation set and use it to reinitialize all K models being trained and 4. keep training with K different $\lambda$'s until end of training. Validation accuracy is improved by using per-layer weight decay with cyclic-learning rate.

While PBO is conceptually similar to [2], it is novel due to the fact that it always picks the best performing checkpoint to initialize all models and continues training them. In contrast, [2] always reinitializes underperforming models. Different layers in a neural network require different degrees of regularization. Using a per-layer weight decay is a novel approach as low-level layers which extract generic features from the input require a lesser degree of regularization compared to high-level layers which are class specific and may contribute to biasing towards common classes, thus requiring a higher degree of regularization. PBO also makes use of Per-Layer Weight Decay which is shown to achieve state-of-the-art on CIFAR100 benchmark.

This paper addresses the problem of class-imbalance in real world data and provides novel approaches to overcome this challenge. This problem is generic in nature that applies to different real world applications and domains. In the context of robotics, class-imbalance can be an issue if the robot needs to recognize a variety of objects and some of these objects are much more common than others. For example, if the robot is designed to recognize objects in a household setting, it is likely to encounter a lot more chairs and tables than rarer objects like violins. If the robot's object recognition model is trained on a dataset that is heavily imbalanced in favor of common objects, it may not perform well on rarer objects. This can be especially problematic if the rarer objects are important for the robot to recognize accurately, such as objects that the robot needs to avoid or objects that are required for a specific task. The methodologies developed in this paper help to provide a generic solution to tackle this problem.

## 4.2 Automatic Detection of Road Work and Construction with Deep Learning Model and a Novel Dataset

Refer [23]. Self-driving cars and autonomous robots are technologies that have been advancing rapidly in recent years. Autonomous vehicles are capable of sensing the environment and navigating without the need for human input. They use a combination of sensors such as cameras, LIDAR, ultrasonic sensors, etc. to gather information about the environment and make decisions about how to navigate. Computer vision techniques greatly find their application in robotic systems, especially in the area of autonomous robotics where these techniques are used for obstacle detection, video object segmentation, vision and language navigation, etc. One of the main goals of an autonomous vehicle (AV) is to ensure safety while navigating in real world environments. The stochastic nature of real world environments renders it difficult to be reliable in the prediction of deep learning models. Even the state-of-the-art methods are unable to interpret and achieve an understanding of the surrounding environment comparable to human beings. For example, road construction or bike lane construction affects the normal, daily transportation which may cause traffic congestion, sometimes even leading to accidents or other hazards. Therefore, it is essential for autonomous vehicles navigating in such real world environments, to detect the presence of construction zones and make decisions that help in alleviating such problems..

This work specifically attempts to address the problems posed by construction zones to the autonomous safe navigation of AVs in the real world. Prior to training any deep learning model, collection of large samples of datasets is a necessity, especially since the state-of-the-art deep learning models require large samples to train and optimize their network parameters. In this work, the authors present a methodology to gather construction zone labeled datasets from both online and offline sources. This work also aims to train a construction identification model and use it to test the dataset and measure its detection accuracy. Further, the authors conduct experiments and analysis to better identity features that are crucial in identifying construction zones. Unavailability of datasets pertaining specifically to construction zones serves as a motivation for this work.

The goal of this work mainly relies on detecting or identifying construction zones that directly affect daily transportation and does not take into account the construction zones that are not on the road. Diversity in a dataset is crucial to prevent overfitting. To this end, the authors suggest a methodology whereby datasets are generated both automatically, using the Google search engine as well as manually, by taking pictures at different locations in the real world. The first method implements a python script to automatically obtain the images along with the labels by specifying query keywords that take the following format

[multiple objectives] + noun

The objectives include weather conditions, location, type of roads, time of the day, etc. while the nouns include keywords such as "road construction", "road word", etc. It was found that, by implementing this method some unsuitable images were also returned such as synthetic images or slides of a presentation and sometimes images with bad scales and unacceptable viewpoints. Therefore to collect a reliable dataset, manual labeling and checking is done so that the undesired data samples are removed.

Although the first method generates diverse datasets, to ensure that a deep learning model to be trained on the dataset can be generalized to real world settings, manual collection of datasets from real world construction zones is adopted. The authors collect datasets by taking pictures in two cities in nearby areas of construction zones so that the distribution of construction and non-construction images are more balanced.

The dataset collected using Google search engine is used for training the model while the datasets collected manually are used to test the model and evaluate its performance in the real world.

The classifier classifies the datasets into construction and non construction zones. The model architecture used in this work builds over the ResNet101 backbone. Fig. 1 shows the different model architectures used in this work.

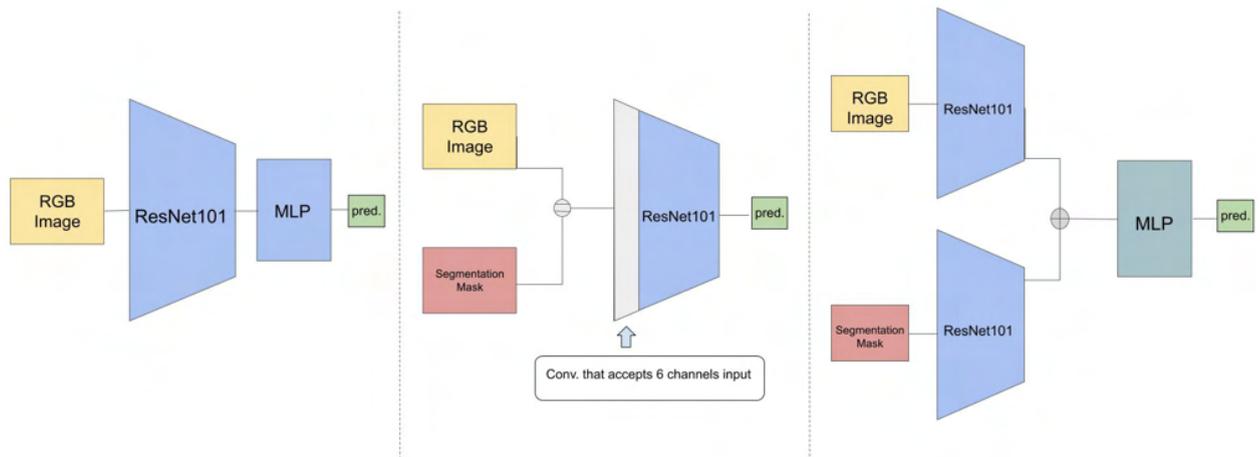

Fig. 1: This figure illustrates the pipeline used for different experiments.

The left pipeline, which is used for basic construction zone classification, accepts a 3 channel RGB image as input. It uses ResNet101 [3] as backbone followed by a multi-layer perceptron(MLP) module. The other 2 architectures shown in the figure, are used when segmentation masks are used as additional feature inputs along with the 3 channel RGB image. The authors hypothesize that the detection accuracy of the model can be improved by using segmentation masks as additional input to the network. The results of the models indicate that the middle architecture, which takes a 6 channel concatenated input of RGB image and segmentation masks, performs poorly as compared to the base architecture at the left of fig1, thus contradicting the hypothesis. However, the two-branch architecture shown in the right of fig1 performs much better than the base model, which indicates that the features of the RGB image and segmentation masks when learnt separately using 2 different ResNet101 backbones provide a higher accuracy of detection than using a single ResNet101 backbone to learn the concatenated features of RGB image and segmentation masks.

The methodology used for data collection ensures that the gathered dataset is diverse and can be trained to accurately identify construction sites. The training data gathered from Google engine by using query keywords can be generalized and adopted to gather data for other deep learning applications such as face recognition, etc. There are no datasets specific for construction zones. This work collects data specifically for the task of identifying construction zones and road work. Therefore, it helps autonomous agents to tackle adverse situations in the real-world, as discussed in this work and enables the deep learning model to generalize better and become more robust in this task.

This work helps spur the development and application of computer vision techniques by generating diverse datasets to tackle the problem of detection of construction zones in the real world. The benefit of this paper is that it demonstrates the utility and effectiveness of the construction zone dataset in a real-world setting.

## 4.3 Semantic Segmentation in Complex Scenes for Robotics Navigation

Refer [24]. Semantic segmentation, the task of assigning semantic labels to each pixel in an image, is a key component in enabling robots to navigate and understand their environment. In complex scenes, however, traditional semantic segmentation methods can struggle to accurately identify and locate objects, due to the presence of occlusions, cluttered backgrounds, and variations in lighting and viewpoint. As a result, autonomous driving becomes more and more challenging in complex and unstructured environments. In general, the autonomous driving environment can be grouped under two categories, on-road and off-road. Most researchers study the on-road and off-road environments separately. For example, Cityscapes [4] dataset specifically covers urban scenes while RUGD [5] covers off-road unstructured scenes. Therefore, due to the limitation of datasets, most models for autonomous navigation achieve high performance in a specific type of environment only.

Moreover, real world environments often require autonomous agents to navigate in complex scenes with both on-road and off-road environments. In an attempt to solve this problem, authors of [6] have proposed to implement semantic segmentation with transfer learning for off-road autonomous driving. Their experiment was successful in detecting obstacles, roads, trees and grass. However, there still remains another challenge of interpreting extreme images, such as too bright or too dark images. To solve the above problems, this paper proposes a lightweight encoder-decoder framework for semantic segmentation. Besides, this paper provides an enhanced method for extreme images.

The proposed architecture can be referred to from Fig. 2.

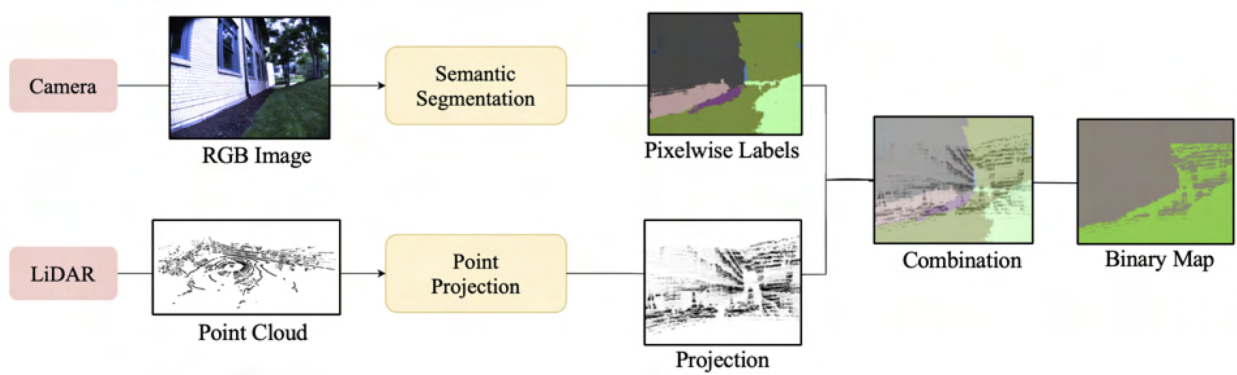

Fig. 2: Overview of the Model.

It consists of two input streams, namely, RGB image and LIDAR point cloud data. Most autonomous vehicles make use of LIDAR data as it provides a 3D representation of the environment which can be used to estimate the distances and positions of objects in the scene. This information can be used to improve the accuracy of semantic segmentation by providing additional context about the location and size of objects in the scene. RGB images, on the other hand, provide a 2D representation of the environment that captures the visual appearance of objects. This information can be used to improve the classification of objects based on their visual appearance. In the proposed model architecture, the RGB image is processed by a lightweight encoder-decoder module such as MobileNetv2 [7], that produces the necessary semantic labels for each pixel. The 3D LIDAR point clouds are projected onto a 2D plane. Both the outputs are combined together to produce a feature map that contains information about semantic labels as well as the height of each object. It is henceforth used to determine traversable and non-traversable regions as shown below in Fig. 3.

Information is retrieved from extreme images by using segmentation masks for uninformative pixels as can be observed from Fig. x. Rectangular blocks are chosen to surround uninformative pixels such that the pixels occupy the centre of the blocks. The most frequent label in each block is assigned to the central pixel and the most frequent label among the pixels is chosen as the class label.

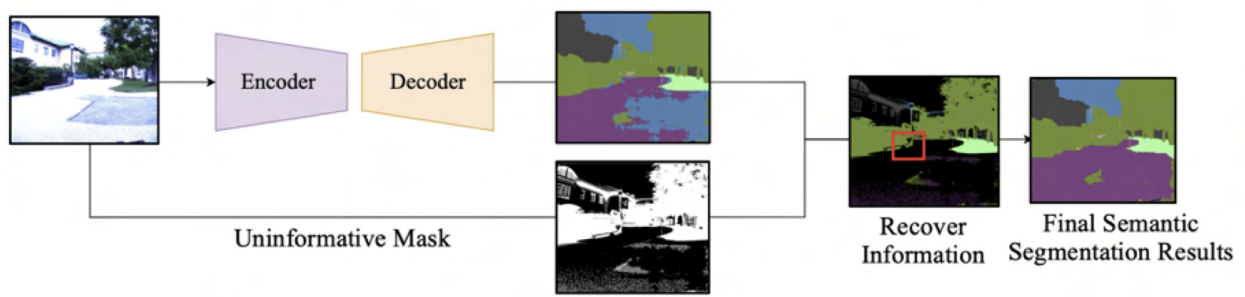

Fig. 3: Semantic Segmentation Framework including tricks on extreme images

The model is experimented on two datasets, Cityscapes and CMU campus frames. The combination results suggest that the proposed architecture performs better than benchmarking results and archives state-of-the-art. Therefore, this paper demonstrates that the proposed solution improves navigation performance for robots. In an effort to further improve the model performance, the future works identified by the authors are: (i) testing the model under different climatic conditions, (ii) collecting dense 3D LIDAR point clouds.

## 4.4 Joint SLAM on Multiple Monocular Cameras for Legged Robots

Refer [25]. Simultaneous Localization and Mapping (SLAM) is a crucial technique for robots to navigate and interact with their environment. It enables robots to build a map of their environment and estimate their location within the map simultaneously. It is essential for various robotic applications such as obstacle detection, autonomous navigation, exploration, etc. SLAM can be implemented using various sensors such as LIDAR, RGB-D cameras, Stereo cameras and Monocular cameras. "Hector SLAM" and "3D SLAM" use data from 2D LIDAR and 3D LIDAR respectively, while "Visual SLAM" or "Camera-based SLAM" use visual data from cameras, to perform simultaneous localization and mapping. SLAM is used in a wide range of robots such as wheeled robots, unmanned aerial vehicles, bipedal or legged robots, etc.

While several SLAM algorithms are shown to be highly effective in case of wheeled robots, it is observed that such techniques perform poorly when applied to systems with rapid and dynamic motion such as legged robots [8]. Legged robots are designed to move in complex and dynamic environments possessing high degrees of freedom. The high degree of freedom of the robot's motion causes violent shaking of on-board sensors of the robot and results in corrupt measurements leading to degraded performance. This paper presents a visual-inertial SLAM method to overcome these challenges in a legged robot.

Typically, the process of visual SLAM involves, (i) Initialization, (ii) Feature detection and tracking, (iii) Motion estimation, (iv) Map updating, (v) Loop closure. The process of visual SLAM is iterative, with the robot or vehicle updating its position and map estimates as it gathers more data. The overall goal of visual SLAM is to provide the robot with a consistent, accurate, and up-to-date map of the environment, as well as an estimate of its own position within that map, even in the absence of GPS signals or other external references. In this work, the authors propose a four-module method, namely, (i) detection, (ii) classification, (iii) tracking, (iv) re-identification, for performing visual SLAM on legged robots.

Fig. 4. shows a simplified example to illustrate the above method.

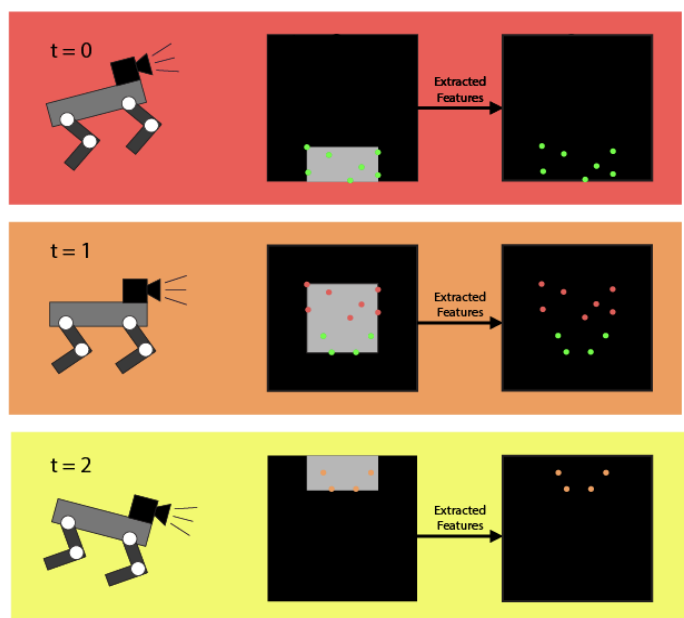

Fig. 4

In the first time-step, features are identified and unique identifications are generated. As the robot's orientation changes in the next time step, the previously detected features (shown in red) are re-identified and new features are added to the map. In the final time-step, previously detected features (shown in orange) are again re-identified. We can infer from Fig. 4 that, the field of vision of the robot changes continuously as the orientation of the robot changes. Therefore, the robot's field of vision can alternatively be demonstrated by representing visual features as objects on a conveyor belt moving between two cameras, as proposed by the authors in this paper. This approach has been demonstrated consistently in many modern SLAM systems. [8][9][10]

Fig. 5 shows the overview of the architecture used.

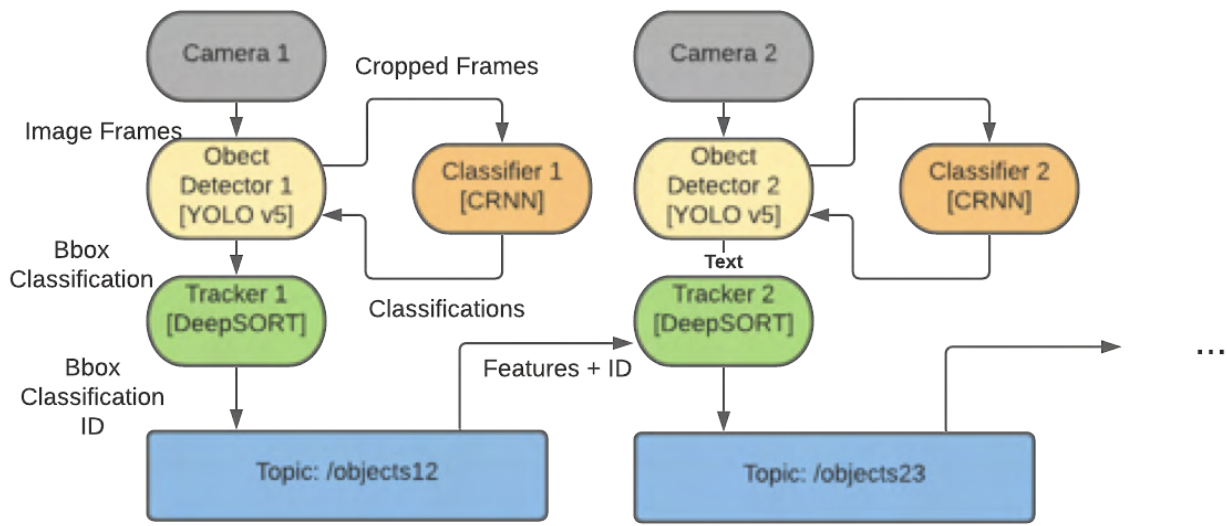

Fig. 5: System diagram for SLAM algorithm.

For the initial detection of objects, this method uses the YOLO v5s network. YOLO v5s has a pre-built classifier and provides the bounding box classifications for objects. The authors modify the original YOLO v5s network by replacing the classifier layer with a RNN based classifier network so that the temporal information from previous detections are preserved. Hence, a joint CNN-RNN(CRNN) network is used to classify cropped images of objects fed into the network at each time step. The classifications and bounding box information is sent to the tracking module to perform multi-object tracking. For this purpose, the DeepSORT network is used to generate a unique ID for each feature so that it can be re-identified between time-steps and between camera views.

The authors perform experiments by integrating the detecting and tracking modules and compare the classification of the CRNN to that of YOLO v5s. They report that the classifications from YOLO v5s are inconsistent for the first few frames while the classifications from CRNN are more smooth and increase consistently until it is almost 1. These results demonstrate the ability of the CRNN to perform more consistent classification than YOLO v5s.

A number of possible future works are suggested in the paper. The experiments conducted in this paper were performed using a single monocular camera. The authors aim to extend this to multiple monocular camera system originally proposed in this paper. The authors also plan to implement this method on a real legged robot and evaluate its performance.

The problem addressed in this paper is generic and can be applied to any system that is susceptible to violent shaking of sensors causing corrupt measurements and poor performance of SLAM algorithms. An active area of investigation could be to develop a LIDAR based SLAM approach to solve the same problem.

## 4.5 Reason & Act : A Modular Approach to Explanation Driven Agents for Vision and Language Navigation

Refer [26]. Vision and language navigation (VLN) refers to the task of navigating an agent through an environment using vision and linguistic cues. The VLN task requires an agent to find a route from an initial state to the target state by following an instruction or a natural language command. At each time-step, the agent must execute an action based on the perceived state of the environment and the natural language instruction. It enables robots or autonomous systems to navigate in real-world environments following natural language instructions. For example, the robot should understand spoken or written instructions such as "go to the hall and turn left" or "go near television and take right" and use the visual features obtained from the environment to infer the meaning of the instruction and navigate to the desired goal. Implementing VLN in real-world settings is a challenging task as it demands autonomous agents to navigate in continuous environments. It involves providing a high-level waypoint to navigate towards by executing a series of low-level actions performed by a local-planner. In an attempt to implement VLN in continuous environments, [11] proposes two end-to-end models that attempt to solve both global planning and low-level control tasks. However, the current methods and models used as a benchmark for performance in VLN-CE have limitations and are not sufficient to handle more complex tasks such as global planning. This motivates the authors to improve the multi-modal understanding of VLN-CE agents by improving high-level navigation. This paper focuses on exploring techniques to improve high-level planning and leaves local planning out of the scope of this work.

Fig. 6. shows the proposed architecture for global planning.

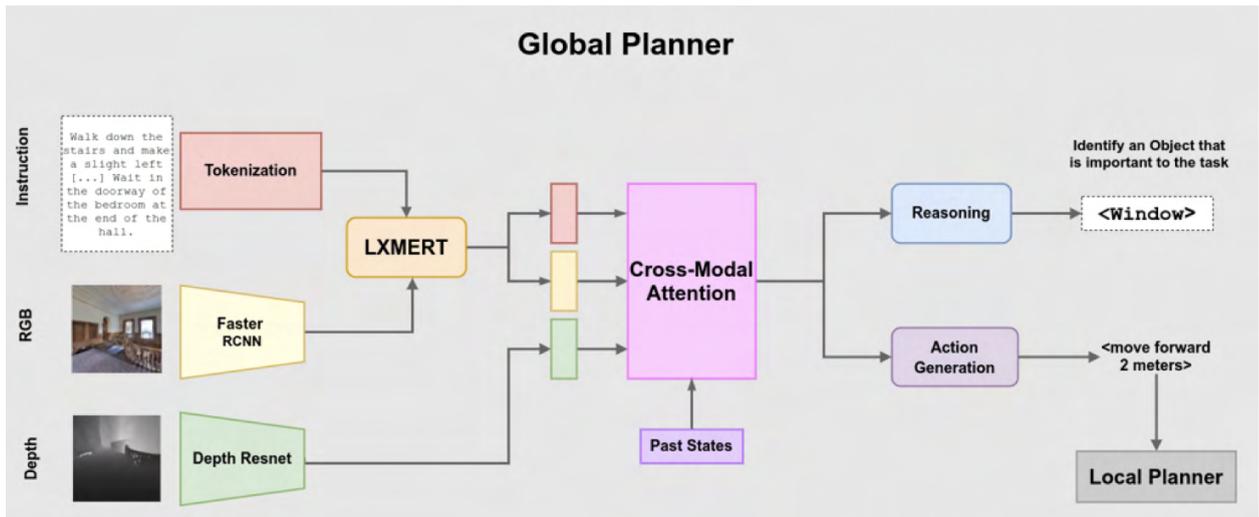

Fig. 6: Global Planner with Cross Modal Attention

The global planner contains two sub-modules, (i) grounding module, (ii)reasoning module. The grounding module creates a strong representation of visual, depth and natural language inputs by finding correlations between the inputs while the reasoning component enhances the multi-modal understanding of the agent by identifying objects in the field of sight that are most important in navigating towards the goal. Pre-trained transformer models are shown to be highly efficient in providing a strong grounded representation of visual and language features. [12][13][14] Drawing inspiration from such works, the authors propose to use LXMERT[15], a transformer based pre-trained encoder, for performing multimodal tasks. The LXMERT model encodes the image features and instruction tokens and performs self-attention and cross-modal attention between the transformer encoder layers. The final layer's output for each image-instruction pair across the vision and language streams of the encoder is extracted and combined to form a representation. Depth observations are passed through a ResNet encoder and feature representations are obtained. The depth features and the combined vision and language features are fused using Cross-Modal Attention(CMA) [12] to create a strong grounded representation. Fig. 7. shows the architecture followed for CMA.

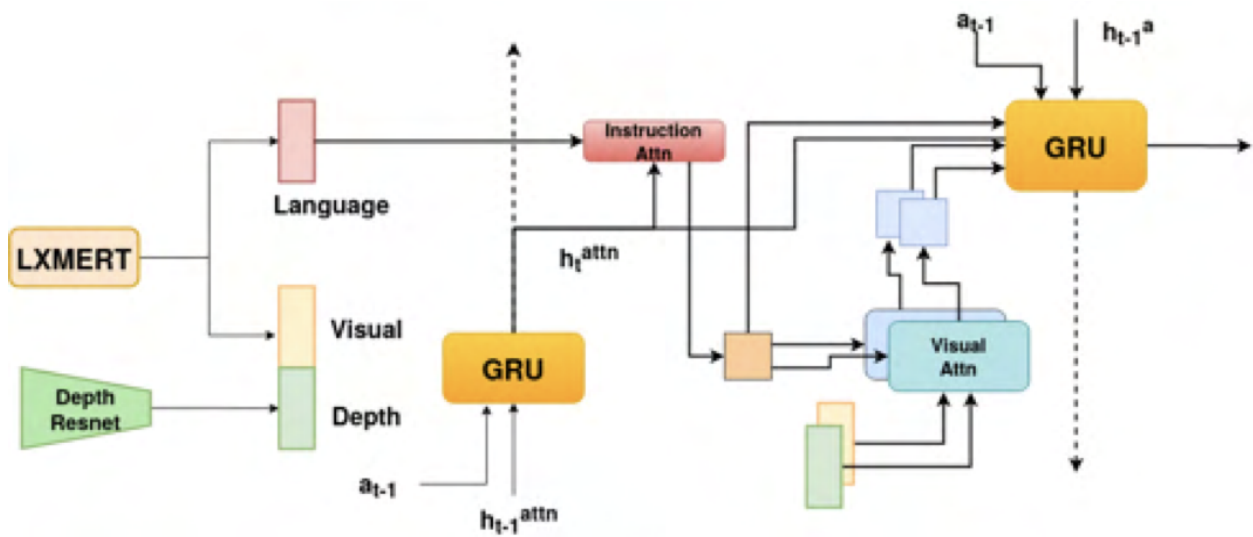

Fig. 7: Cross Modal Attention

The grounded representation is used for action generation as well as for reasoning. For implementing the reasoning component, the grounded features are passed through a linear layer followed by a Softmax activation function. The probability distribution outputs the most crucial object in the field of sight of the agent.

The authors train the global agent using AI Habitat [16] and the VLN-CE dataset [12]. Experiments are performed using this dataset to compare the performance of the global agent against the baseline models for Reasoning and the Vision- Language grounding modules. It was found that, in spite of using LXMERT which contains much more parameters than baseline models, comparable results were achieved. In an effort to improve the performance, the authors suggest replacing FasterRCNN from the LXMERT pipeline with a more efficient feature extractor.

The authors propose several future works. They aim to explore better alternatives to perform the reasoning module. They also plan on integrating the high-level planner with a local planner and test its performance.

The VLN problem chosen in this paper is a challenging task that requires a combination of techniques from computer vision, natural language processing, and robotics. This could be used in a variety of applications, such as service robots, home automation, and search-and-rescue robots. An active area of investigation could be to apply the proposed method in real-world settings and gather datasets using a real robot. This opens up the possibility of better performance of the model in the real-world as compared to simulation environments.

## 4.6. Transfer Exploration in RL: A Study on Recent Count-Based Methods

Refer [18]. The learning process in RL is based on a feedback system in the form of rewards. The agent performs an action on the environment given the state and receives a reward and the next state as a feedback from the environment. The design of the reward system is arbitrary and depends on the application on hand. For some applications, such as Atari games, where the effectiveness of RL was demonstrated as a major machine learning methodology, the rewards are immediate for each action step. However, for many other applications including those that are core problems in Robotics, the rewards can be rare. This makes the learning hard or sometimes not viable. Task agnostic exploration in reinforcement learning is an important research area, especially in settings with sparse or unknown rewards. Learning efficient exploration in these settings remains a key challenge.

Several approaches to address such problems of sparse rewards are found in the literature. The paper [4] introduces Count-Based Exploration Transfer (C-BET) and proposes a two-step transfer based approach. Here the agent first learns to explore across many environments without any extrinsic goal in a task agnostic manner and then it transfers the learned policy to better explore new environments when solving tasks. The issue of sparse rewards has been addressed in step 1 by means of task agnostic exploration. Earlier works on task agnostic exploration were based on intrinsically motivated rewards, such as prediction curiosity, empowerment or visitation counts that defined such rewards in an agent-centric manner.

Authors of C-BET argue that apart from an 'agent-centric' component, there is an 'environment-centric' component to exploration, which can be learned from prior knowledge and experiences. The agent-centric component can model behaviours like inherent curiosity and this may lead the agent to seek out surprises. Imagine a human sitting in the same place, observing the same room and objects. This leads to boredom and may lead him to explore unseen places. Likewise, the agent may find it more interesting when it observes high-impact objects or activities associated with the environment. For instance, imagine a human who observes a bookshelf in a room. He may find it interesting to look at it in detail, particularly if he is a voracious reader. Thus, exploring in a task agnostic scenario could be thought of as due to both an inherent curiosity element of an agent and also interesting objects in the environment. The design of a task agnostic rewards system could be designed following the above model. The authors of the original paper on C-BET report that their technique (a) learns more effectively when placed in a multi-environment setup, and (b) either outperforms or performs competitively with prior methods across several unseen testing environments.

In Fig 8 we show the two step process of transfer learning that is shown to address the problem of sparse rewards using task agnostic learning as a pre-training step followed by task specific learning.

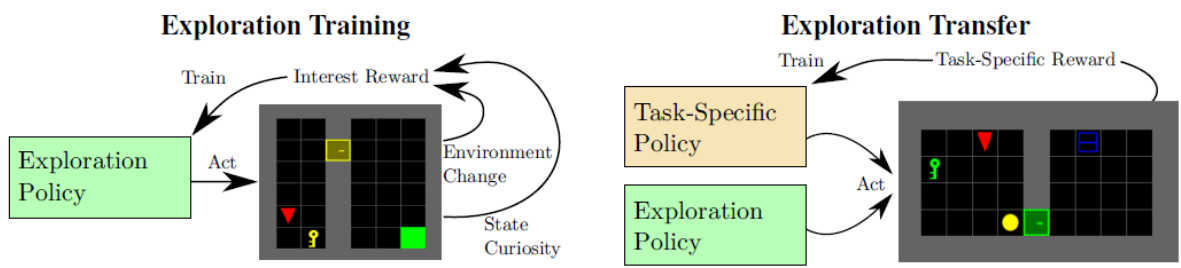

Fig 8: C-BET Pre-training and C-BET Transfer

Jacob Adkins et al conducted a study on this base work of C-BET by extending this technique and reported in their paper [20]. Here, the authors propose an improvement over C-BET to improve over the exploration performance. Specifically, the investigation is based on the question: what happens when an agent not only uses a pre-trained exploration policy but continues relying on intrinsic motivation while learning a task as well? This agent will attempt to both solve its extrinsic task and maintain curiosity about the world during the stage 2 of transfer learning. The goal of C-BET as well as this proposed extension is to learn exploration policies that encourage interaction with multiple environments and transfer well to new environments, i.e., that can further be trained to solve extrinsic tasks faster. Besides this, the authors also believe that this extension might help not only the current task on hand but also any related future tasks by using intrinsic motivation during the task specific exploration.

Experiments were conducted on both Minigrid environments [17] as well as on Habitat 2.0 [19], which are simulation environments. While Minigrid simulates a set of grid-like states, Habitat enables more diverse and realistic simulation where the agent can navigate through visually realistic rooms.

The authors report, for the grid environment, transferring the state change counts does not improve performance on the extrinsic task nor exploration as measured by interactions per episode. The experiments and further analysis for the Habitat 2.0 is in early stages.

A number of possible future works are suggested in the paper. One of them is to extend the C-BET to 7 dimensional continuous action space with the discrete grasping action. Another task that the authors propose to do is to test the technique against the Home Assistance Benchmark, which is the suite of higher level tasks and actions.

The problem addressed by C-BET and its extensions is generic and is applicable to a number of applications. Sample efficient exploration and intrinsic motivation are crucial to successful robot learning in the real world. An active area of investigation could be to evaluate sample efficient exploration in real-world settings.

## 4.7. Multi-agent Hierarchical Reinforcement Learning in Urban and Search Rescue

Refer [27]. This paper deals with an important application problem that requires a multi-agent solution. In an urban search-and-rescue (USAR) task, a team of robots are engaged in rescue based activities such as locating victims in different rooms in the environment, clear rubble, and triage victims. In this problem also we encounter an instance of sparse rewards.

In a large campus with many buildings and rooms, it is difficult to have prior information on the exact number of victims who might be trapped and their locations. The agent gets the reward only when it is able to rescue a victim and may get a negative reward when it misses anyone. That is, because the objective of the team is to rescue victims, agents can only receive extrinsic rewards when a victim is successfully triaged, which is a rare event. As these events are rare, the rewards are sparse and delayed. The spatial extent of the environments encountered during USAR missions are typically large and hence may involve a large state space.

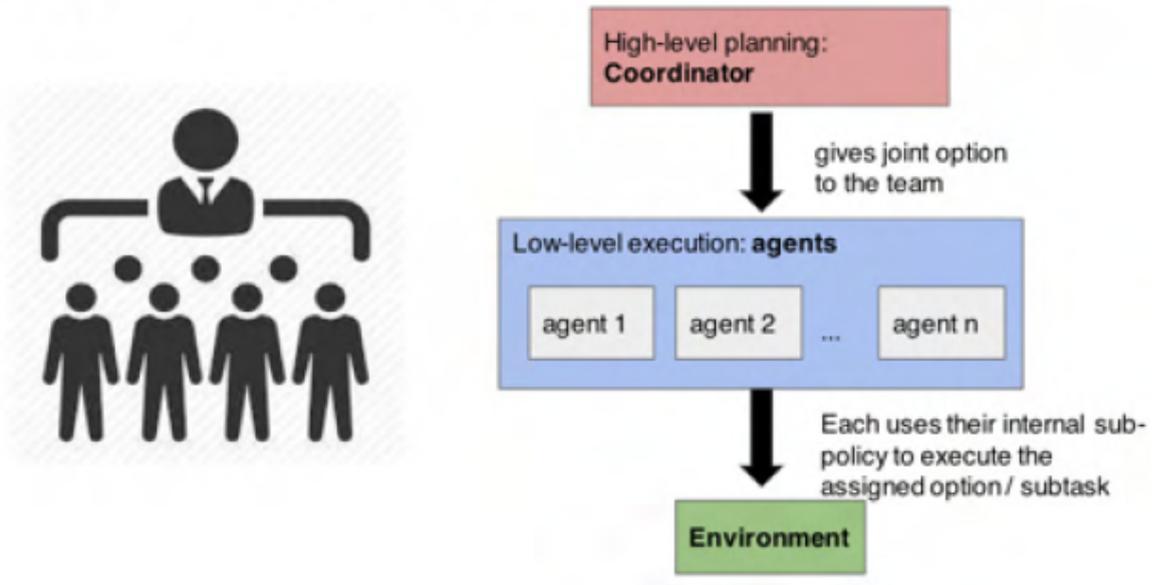

Fig 9: Coordinator-Agent Hierarchical Framework

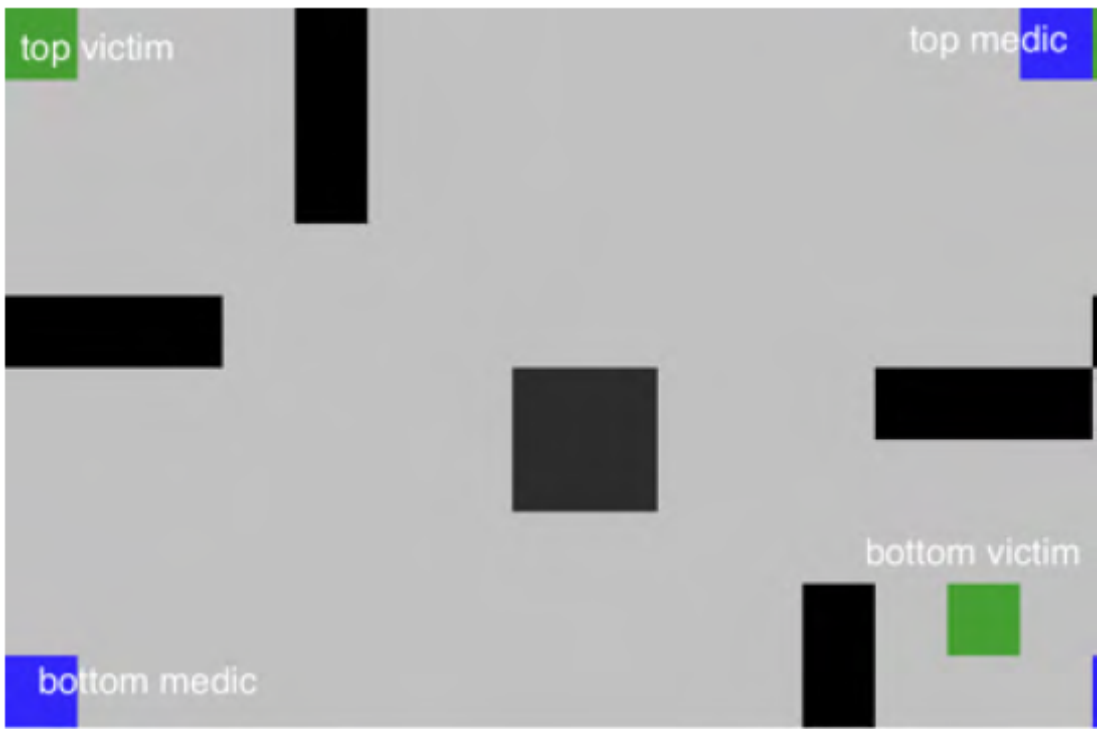

Fig 10: The two-room environment. The walls are in black. Two green squares are two victims. Two blue squares are two agents

Given these characteristics and the distributed nature of this application domain, the authors propose a solution based on a team of robotic agents cooperating with each other. These agents are coordinated using a coordinator algorithm, following a hierarchical multi-agent pattern as shown in Fig 9. An example of a 2 room environment and the representation of agents and victims is shown in Fig 10.

To deal with this hard-exploration and large state space, hierarchical approaches both over action and state spaces are proposed. In this model, each low level agent is assigned a set of sub tasks and trained to execute in a completely decentralized manner using low level features of subsets of state space. The agents are controlled by the coordinator that operates at the higher level and trained on a high level graph representation. The coordinator determines the tasks to be assigned to the agents and issues the suitable commands. For example, "Agent 1 goes to search for victim from location A, Agent 2 for victim from location B" and so on. Each agent performs the lower level tasks as required by the coordinator and reports the rewards back to complete the loop. By constraining a given agent to plan over a smaller subset of the environment, the state space handled by a given agent is made smaller and the coordinator plans over a small subgraph of the environment for the agent. This architecture also mitigates the long horizon problem where the sequences are too long based on the size of the environment and the difficulty in locating the victims.

The architecture proposed in this paper draws inspiration from two popular frameworks: Options by Sutton et al [30] and Feudal learning [31], [32]. The Options approach introduces the concept of "macro actions", that are high level actions that are further decomposed to low level actions. The Feudal approach imposes a hierarchy where there are managers (lords) and sub managers (serfs). As we saw, the approach followed combines the key ideas from both these frameworks.

In the two room environment example mentioned above, the agents are tasked with low level actions while the coordinator performs the macro-actions as stated in the Options framework. The distribution of these tasks in a hierarchical structure, where the coordinator acts as the manager and agents as sub managers follow the Feudal framework approach. The coordinator deals with abstract spaces and actions. The spaces are represented by graphs, see Fig 11 for an illustration.

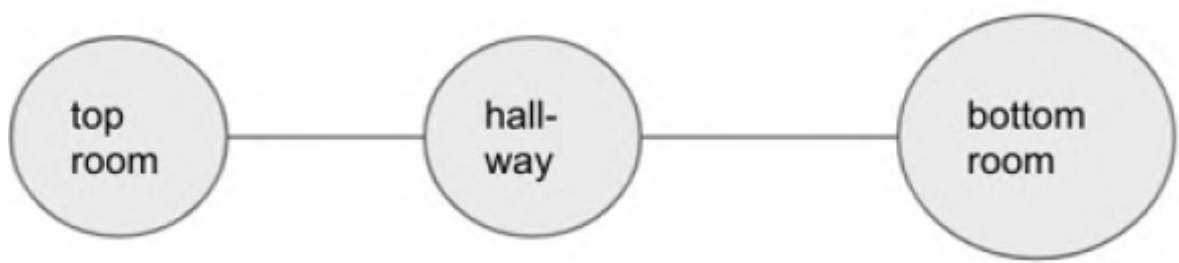

Fig 11: The graph representation for the two-room environment.

The authors have implemented both the training of sub-policies and the coordinator's meta-policy using Deep Q networks with experience replay and target nets [33]. Some of the future work are: extending this architecture to environments of larger size, with heterogeneous agents of different capabilities, using Graph Neural Networks (GNN) to encode the environment information and formulate the Reinforcement Learning problem as a node-level decision-making task. The authors also have identified other techniques such as transfer learning and Object-oriented model-based Reinforcement Learning for future investigation.

## 5. Conclusion and Future Work

We surveyed over 15 papers from RISS, CMU Journal 2021 and presented a sample of 7 papers in this survey. We identified possible extensions that might be good candidates for consideration for internship. In future, we plan to cover a larger subset of papers, widen the scope to include relevant research from other literature and turn this draft report into a formal paper. We also plan to run some of these projects on a physical robot and investigate the challenges in moving from a simulated environment into the real world scenario. We also plan to implement some ideas we discovered from this survey effort.